\documentclass{bmvc2k}

\usepackage{times}
\usepackage{epsfig}
\usepackage{graphicx}
\usepackage{amsmath}
\usepackage{amssymb}
\usepackage{color}
\usepackage{colortbl}
\usepackage{algorithmicx,algorithm}
\usepackage{multirow}
\usepackage{subfigure}
\usepackage{bm}
\usepackage{ctable}
\definecolor{Gray}{gray}{0.9}
\definecolor{White}{gray}{1}

\title{HandTailor: Towards High-Precision Monocular 3D Hand Recovery}

\addauthor{Jun Lv}{lyujune_sjtu@sjtu.edu.cn}{12}
\addauthor{Wenqiang Xu}{vinjohn@sjtu.edu.cn}{1}
\addauthor{Lixin Yang}{siriusyang@sjtu.edu.cn}{1}
\addauthor{Sucheng Qian}{qiansucheng@sjtu.edu.cn}{1}
\addauthor{Chongzhao Mao}{chongzhao.mao@flexiv.com}{2}
\addauthor{Cewu Lu}{lucewu@sjtu.edu.cn}{1}

\addinstitution{
 Department of Computer Science\\
 Shanghai Jiao Tong University\\
 Shanghai CHINA
}
\addinstitution{
 Flexiv Ltd.\\
 Shanghai CHINA
}

\runninghead{Jun Lv ET AL.}{HandTailor}

\begin{document}

\maketitle

\begin{abstract}
3D hand pose estimation and shape recovery are challenging tasks in computer vision. We introduce a novel framework HandTailor, which combines a learning-based \textit{hand module} and an optimization-based \textit{tailor module} to achieve high-precision hand mesh recovery from a monocular RGB image. The proposed \textit{hand module} adapts both perspective projection and weak perspective projection in a single network towards accuracy oriented and in-the-wild scenarios. The proposed \textit{tailor module} then utilizes the coarsely reconstructed mesh model provided by the \textit{hand module} as initialization to obtain better results. The \textit{tailor module} is time-efficient, costs only $\sim 8ms$ per frame on a modern CPU. We demonstrate that HandTailor can get state-of-the-art performance on several public benchmarks, with impressive qualitative results. Code and video are available on our project webpage~\href{https://sites.google.com/view/handtailor}{https://sites.google.com/view/handtailor}.
\end{abstract}

\section{Introduction}
The hand is one of the most important elements of humans when interacting with the environment. Single image 3D hand reconstruction is a task that seeks to estimate the hand model from a monocular RGB image, which could be beneficial for various applications like human behavior understanding, VR/AR, and human-robot interaction.

Recovery of the 3D hand model from a single image has been studied for decades, but challenges remain. First, since the 3D hand reconstruction is a process of 2D-3D mapping, the learning method should somehow handle the camera projection. Some of the previous works \cite{zhou2020monocular,kulon2020weakly} choose to ignore this issue by only predicting root-relative hand mesh, which limits the application of their algorithms. Others rely on perspective projection \cite{iqbal2018hand,yang2020bihand} or weak perspective projection \cite{boukhayma20193d,baek2019pushing,zhang2019end}. Perspective projection is more accurate but requires intrinsic camera parameters, making it impossible to apply without camera information. While weak perspective projection can be applied to in-the-wild cases but the approximation is conditioned. For previous works, once the camera projection model is selected, the adaptability of the method is also determined. Second, predicting 3D mesh usually cannot guarantee back-projection consistency, which means, a visual appealing predicted 3D hand may have numerous evident errors, such as a few degrees of finger deviation, when it is re-projected onto the original image (See Fig. \ref{fig:quantitative}).


To address these two issues, we propose a novel framework named \textbf{HandTailor}, which consists of a CNN-based hand mesh generation module (\textit{hand module}) and an optimization-based tailoring module (\textit{tailor module}). The \textit{hand module} is compatible with both perspective projection and weak perspective projection without any modification of the structure and model parameters. It can be used to project the 3D hand more accurately when the camera parameters are available and also can be used to predict the in-the-wild image by simply changing the computational scheme. The \textit{tailor module} can refine the rough hand mesh predicted by \textit{hand module} to higher precision and fix the 2D-3D miss-alignment based on more reliable intermediate results.
With the initialization provided by the \textit{hand module} and the differentiability of the \textit{tailor module}, the optimization adds only $\sim 8 ms$ overhead.

Experiments show that HandTailor can achieve comparable results with several state-of-the-art methods, and can accomplish both perspective projection and weak perspective projection. The \textit{tailor module} can improve the performance quantitatively and qualitatively. On a stricter $AUC_{5-20}$ metric, HandTailor gets 0.658 with an improvement close to 0.1 by the \textit{tailor module} on RHD \cite{zimmermann2017learning}. To prove the applicability and generality of the \textit{tailor module} upon other intermediate representation-based approaches \cite{yang2020bihand,boukhayma20193d,zhou2020monocular}, we run several plug-and-play tests, which also show performance improvements by a large margin.

Our contributions can be summarized as follows: First, we propose a novel framework HandTailor for single image 3D hand recovery task, which combines a learning-based \textit{hand module} and an optimization-based \textit{tailor module}. The proposed HandTailor achieves state-of-the-art results among many benchmarks. Second, this method adapts both weak perspective projection and perspective projection without modification of the structure itself. Such architecture can be applied to both in-the-wild and accuracy-oriented occasions. Third, the \textit{tailor module} refines the regressed hand mesh by optimizing the energy function w.r.t the intermediate outputs from the \textit{hand module}. It can also be used as an off-the-shelf plugin for other intermediate representation-based methods. It shows significant improvements both qualitatively and quantitatively.

\section{Related Works}
In this section, we discuss the existing 3D hand pose estimation and shape recovery methods. There are lots of approaches based on depth maps or point clouds data \cite{ge2016robust,yuan2018depth,malik2020handvoxnet,ge2018hand,mueller2017real,wan2020dual}, but in this paper, we mainly focus on single RGB-based approaches.

\noindent{\bf 3D Hand Pose Estimation.} Zimmermann and Brox \cite{zimmermann2017learning} propose a neural network to estimate 3D hand pose from a single RGB image, which lays the foundation for subsequent research. Iqbal et al. \cite{iqbal2018hand} utilize a 2.5D pose representation for 3D pose estimation, provide another solution for 2D-3D mapping. Cai et al. \cite{cai2018weakly} propose a weakly supervised method by generating depth maps from predicted 3D pose to gain 3D supervision, which gets rid of 3D annotations. Mueller et al. \cite{mueller2018ganerated} use CycleGAN \cite{CycleGAN2017} to bridge the gap between synthetic and real data to enhance training. Some other works \cite{spurr2018cross,gu20203d,yang2019aligning,yang2019disentangling,zhao2020knowledge} formulate 3D hand pose estimation as a cross-modal problem, trying to learn a unified latent space. 

\noindent{\bf 3D Hand Mesh Recovery.} 3D mesh is a richer representation of human hand than 3D skeleton. To recover 3D hand mesh from monocular RGB images, the most common way is to predict the parameters of a predefined parametric hand model like MANO \cite{romero2017embodied}. Boukhayma et al. \cite{boukhayma20193d} directly regress the MANO parameters via a neural network and utilize a weak perspective projection to enable the in-the-wild scenes. Baek et al. \cite{baek2019pushing} and Zhang et al. \cite{zhang2019end} utilize a differential renderer \cite{kato2018neural} to gain more supervision from hand segmentation masks. These methods generally predict the PCA components of MANO parameters, causing inevitable information loss. To address this issue, Zhou et al. \cite{zhou2020monocular} propose IKNet to directly estimate the rotations of all hand joints from 3D hand skeleton. Yang et al. \cite{yang2020bihand} reconstruct hand mesh with multi-stage bisected hourglass networks. Chen et al. \cite{chen2021model} achieve camera-space hand mesh recovery via semantic aggregation and adaptive registration. Different from the aforementioned model-based method, there are also some approaches \cite{ge20193d,kulon2019single,kulon2020weakly} that generate hand mesh through GCN \cite{defferrard2016convolutional}, providing new thoughts for this task. \cite{zimmermann2021contrastive,chen2021model} try to accomplish this task with self-supervise learning. Different from the aforementioned method, we propose a novel framework that combines a learning-based module and an optimization-based module to achieve better performance, and also adapts both weak perspective projection and weak perspective projection for high precision and in-the-wild scenarios.

\noindent{\bf Optimization-based 3D Hand Mesh Recovery.} Apart from the learning-based methods, there are also some other attempts on reconstructing hand mesh in an optimization-based manner. Previous works choose to fit the predefined hand model \cite{oikonomidis2011efficient,tkach2016sphere} to depth maps \cite{tkach2016sphere,oikonomidis2011efficient,tagliasacchi2015robust,sharp2015accurate}. For monocular RGB reconstruction, Panteleris et al. \cite{panteleris2018using} propose to fit a parametric hand mesh onto 2D keypoints extracted from RGB image via a neural network \cite{cao2017realtime}. Mueller et al. \cite{mueller2018ganerated} introduce more constraints for better optimization, like 3D joints locations. Kulon et al. \cite{kulon2020weakly} utilize iterative model fitting to generate 3D annotations from 2D skeleton to achieve weakly supervise learning, treating the recovery result of the optimization-based method as the upper bound of the learning-based method. Though these optimization-based methods share a similar ideology to our \textit{tailor module}, 
our approach exploits the multi-stage design of the \textit{hand module} to make use of information from different stages and accelerates the optimization process. The reduced overhead makes the optimization possible while in inference, which is crucial for practical usage.

\section{Method} 

\subsection{Preliminary}
\noindent{\bf MANO Hand Model.} MANO \cite{romero2017embodied} is a kind of parametric hand model, which factors a full hand mesh to the pose parameters $\theta \in \mathbb{R}^{16 \times 3}$ and the shape parameters $\beta \in \mathbb{R}^{10}$. 
The hand mesh $\mathcal{M}(\theta, \beta) \in \mathbb{R}^{V \times 3}$ can be obtained via a linear blend skinning function $\mathcal{W}$,
\begin{equation}
    \mathcal{M}(\theta,\beta)=\mathcal{W}(\mathcal{T}(\beta,\theta),\mathcal{J}(\beta),\theta,\omega)
\end{equation}
$\mathcal{T}$ is the rigged template hand mesh with 16 joints $\mathcal{J}$. $\omega$ denotes the blend weights, and $V=778$ is the vertex number of hand mesh. For more details please refer to \cite{romero2017embodied}. 

\noindent{\bf Camera Models.} 
Perspective projection describes the imaging behavior of cameras and human eyes. To transform 3D points in camera coordinate to 2D pixels in image plane, we need camera intrinsic matrix $\mathcal{K} \in \mathbb{R}^{3 \times 3}$,
\begin{equation}
   \begin{bmatrix} u \\ v \\ 1 \end{bmatrix} = \pi(\mathcal{K};  \begin{bmatrix} x \\ y \\ z \end{bmatrix}) = \frac{1}{z} \begin{bmatrix} f_x & 0 & c_x \\ 0 & f_y & c_y \\ 0 & 0 & 1 \end{bmatrix} \begin{bmatrix} x \\ y \\ z \end{bmatrix}
\label{equ:perspective}
\end{equation}
$\pi(\cdot)$ is the projection function, ($f_x$, $f_y$) are focal lengths, and ($c_x$, $c_y$) are camera centers.

Weak perspective projection takes a simplification on the intrinsic matrix, formulated as:
\begin{equation}
   \begin{bmatrix} u \\ v \\ 1 \end{bmatrix} = \Pi(\mathcal{K}^\prime ; \begin{bmatrix} x \\ y \\ z \end{bmatrix}) = \frac{1}{z} \begin{bmatrix} s & 0 & 0 \\ 0 & s & 0 \\ 0 & 0 & 1 \end{bmatrix} \begin{bmatrix} x \\ y \\ z \end{bmatrix} 
\label{equ:weakperspective}
\end{equation}
$\Pi(\cdot)$ is the weak perspective projection function, $s \in \mathbb{R}$ is the scaling factor. To align the re-projected model to the image, we also need camera extrinsic parameters for rotation $\mathcal{R} \in \mathbb{R}^{3 \times 3}$ and translation $t \in \mathbb{R}^{3}$. In practice, we set $\mathcal{R}$ as identity matrix and predict $t$ solely.

\subsection{Overview}

\begin{figure*}[t]
   \begin{center}
   \includegraphics[width=0.85\linewidth]{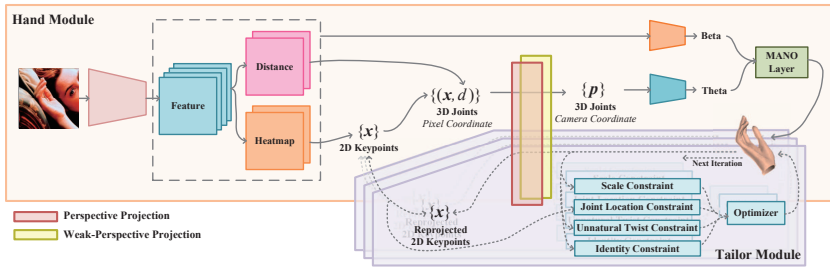}
   \end{center}
   \vspace{-5mm}
      \caption{HandTailor consists of two components, \textit{hand module} and \textit{tailor module}. \textit{Hand module} predicts the parameters of MANO through the pose and shape branches to generate hand mesh. It takes a multi-stage design and produces 2D keypoints and 3D joints as intermediate results. The \textit{tailor module} optimizes the predicted hand mesh according to several constraints. HandTailor adapts both perspective projection and weak perspective projection.}
\label{fig:pipeline}
\vspace{-5mm}
\end{figure*}

The proposed HandTailor consists of two components, a learning-based \textit{hand module} (Sec. \ref{sec:handmodule}) and an optimization-based \textit{tailor module} (Sec. \ref{sec:tailormodule}), which aims to reconstruct the 3D hand mesh from RGB images. The overall pipeline is shown in Fig. \ref{fig:pipeline}.

\subsection{Hand Module} 
\label{sec:handmodule}

\noindent{\bf Intermediate Representation Generation.} Given RGB image $\mathcal{I} \in \mathbb{R}^{H \times W \times 3}$ 
we utilize a stacked hourglass network \cite{newell2016stacked} to generate feature space $\mathcal{F} \in \mathbb{R}^{N \times H \times W}$.
Then the 2D keypoint heatmaps $\mathcal{H} \in \mathbb{R}^{k \times H \times W}$, and distance maps $\mathcal{D} \in \mathbb{R}^{k \times H \times W}$ are predicted from $\mathcal{F}$. The sum of $\mathcal{H}$ is normalized to 1 at each channel, $\mathcal{D}$ is the root-relative and scale-normalized distance for each joint. We 
scale the length of the reference bone to 1. 

\noindent{\bf Pose-Branch.} This branch is to predict $\theta$. We first retrieve 2D keypoints $\mathcal{X} =\{\bm{x}_i|\bm{x}_i=(u_i, v_i)\in \mathbb{R}^{2}\}^k_{i=1}$ from heatmap $\mathcal{H}$. For $j^{\rm{th}}$ joint, the 2D keypoint $\bm{x}_j \in \mathcal{X}$ and the root-relative scale-normalized depth $d_j$ of $j^{\rm{th}}$ joint in the image plane can be achieved by 
\begin{equation}
    (\bm{x_j}, d_j)^\top = (\sum_{\bm{x}\in\mathcal{H}} {H}^{(j)}(\bm{x}) \cdot \bm{x}^\top, \sum_{\bm{x}\in\mathcal{H}} {H}^{(j)}(\bm{x}) \cdot D^{(j)}(\bm{x}))
\end{equation}
$(\bm{x_j}, d_j)=[u_j, \ v_j, \ d_j]$ is the scaled pixel coordinate of $j^{\rm{th}}$ joint. The root-relative and scale-normalized depth $d_j$ are friendly for neural network training.
To obtain the 3D keypoints $\mathcal{P} =\{\bm{p}_i|\bm{p}_i=(x_i, y_i, z_i)\in \mathbb{R}^{3}\}^k_{i=1}$, we project the $[u_j, \ v_j, \ d_j]$ into camera coordinate as $\bm{p}_j = [x_j, y_j, z_j] \in \mathcal{P}$. To do so, first we recover the depth of each joint by predicting the root joint depth $d_{root} \in \mathbb{R}$. For $j^{\rm{th}}$ joint, the scaled pixel coordinate is converted to
\begin{equation}\label{equ:intri1}
    \bm{p}_j^\top = \pi^{-1}(\mathcal{K}, (\bm{x}_j, d_j)^\top + (\bm{0}, d_{root})^\top)
\end{equation}
The $d_{root}$ is also scale-normalized to comply with the projection model. $\pi^{-1}(\cdot)$ is the inverse function of $\pi(\cdot)$.
Till now, we have the joint locations in camera coordinate.
Following \cite{zhou2020monocular}, we train an IKNet to predict the quaternion of each joint, which is pose parameter $\theta$.

For better convergence, we attach each transformation a loss function. $\mathcal{L}_{kpt^{2D}}$ is defined as the pixel-wise mean squared error (MSE) between the prediction and ground-truth $\mathcal{H}$. $\mathcal{L}_{kpt^{3D}}$ is the MSE between the prediction and ground-truth $\mathcal{P}$. $\mathcal{L}_{d}$ measures the MSE between the prediction and ground-truth $d_{root}$. The $\lambda$s are the coefficients for each term. 
\begin{equation}
    \mathcal{L}_{\theta} = 
    \lambda_{kpt^{2D}}\mathcal{L}_{kpt^{2D}} + \lambda_{kpt^{3D}}\mathcal{L}_{kpt^{3D}}  + \lambda_{d}\mathcal{L}_{d}
\label{equ:losspose}
\end{equation}
To train IKNet, we adopt the same formulation as in \cite{zhou2020monocular}. Note that IKNet needs to be pretrained with $\mathcal{L}_{ik}$, but it also needs to be fine-tuned during the end-to-end training stage.

\noindent{\bf Shape-Branch.} The shape-branch is to predict the shape parameter $\beta$, which takes $\mathcal{F}$, $\mathcal{H}$, and $\mathcal{D}$ as inputs via several ResNet layers \cite{he2016deep}. Though we cannot directly supervise the $\beta$, the network can learn it from indirect supervision, such as re-projection to silhouette or depth, as discussed in Sec. \ref{sec:ablation}. However, in practice, we find a simple regularization $\mathcal{L}_{\beta} = \|\beta\|^2$ is good enough for both training speed and final performance.

\noindent{\bf Mesh Formation.} With $\theta$ and $\beta$, we can finally generate the mesh $\mathcal{M}$ through MANO layer \cite{romero2017embodied}. The MANO layer cancels the translation, thus we need to add it back by translating the root location of $\mathcal{M}$ to $\bm{p}_{root} \in \mathcal{P}$, which is the 3D location of the root joint, to obtain $\hat{\mathcal{M}}$.
The network should also be trained with a $\mathcal{L}_{mano}$, which is a function of both $\theta$ and $\beta$, measures the MSE between ground-truth and the predicted 3D joints $\mathcal{P}_{mano} \in \mathbb{R}^{k \times 3}$ extracted from $\mathcal{M}$. 

\noindent{\bf Overall Loss Function.} Then the overall loss function is
\begin{equation}
    \mathcal{L} = 
    \lambda_{\theta}\mathcal{L}_{\theta} + 
    \lambda_{\beta}\mathcal{L}_{\beta} + \lambda_{mano}\mathcal{L}_{mano}
\label{equ:lossoverall}
\end{equation}
The $\lambda$s are the coefficients for each term with corresponding subscripts.

\subsection{Tailor Module} 
\label{sec:tailormodule}
Current multi-stage pipelines for hand mesh reconstruction \cite{yang2019disentangling,zhou2020monocular}, including ours, which usually construct the early stage for 2D information prediction and 3D information in the later stage, suffer from a precision decrease along the stages. A major reason behind this is that 2D information prediction is less ill-posed than 2D-3D mapping problem, which requires less but actually has more high-quality training samples. The \textit{tailor module} is designed upon this observation to refine the hand mesh $\mathcal{M}$. It compares the output mesh to the intermediate representations from the multi-stage neural network. Since 2D keypoint is the most widely adopted representation and considered to be the most accurate one, we will discuss how to optimize with it. As for other intermediate representations, please refer to Sec. \ref{sec:ablation}. 

To fit the $\hat{\mathcal{M}}$ with the 2D image plane correctly and obtain a better hand mesh $\mathcal{M}^*$, we need to handle three constraints, namely hand scale, joint locations, and unnatural twist.

\noindent{\bf Scale Constraint.} The predicted $\hat{\mathcal{M}}$ may appear inconsistent scale when re-projecting to the image, due to regression noise. 
We can optimize scale-compensate factor $s^* \in \mathbb{R}$ with an energy function, which leads to a more reasonable scale when projecting to the image plane.
\begin{equation}
\label{equ:scaleconstrain}
    \mathcal{E}_s(s^*) = 
    ||\pi(\mathcal{K};s^*\mathcal{P}_{mano}(\theta, \beta)+\bm{p}_{root})-\mathcal{X}||^2_2
\end{equation}

\noindent{\bf Joint Location Constraint.} As mentioned earlier, 2D keypoint estimation usually has higher accuracy. Thus when a well predicted $\mathcal{P}_{mano}$ is projected back to the image plane, it should be very close to the predicted $\mathcal{X}$.
\begin{equation}
\label{equ:jointconstrain}
    \mathcal{E}_J(\theta,\beta) = 
    ||\pi(\mathcal{K};s^*\mathcal{P}_{mano}(\theta, \beta)+\bm{p}_{root})-\mathcal{X}||^2_2
\end{equation}

\noindent{\bf Unnatural Twist Constraint.} Since the losses are mostly joint location-oriented, it is very likely to cause a monster hand.
We follow the design of \cite{zhang2019end} to repair such monster hand pose. Let $\bm{p}_a$, $\bm{p}_b$, $\bm{p}_c$, $\bm{p}_d \in \mathcal{P}_{mano}$ denote 4 joints of the fingers in tip to palm order. $\vec{V}_{ab}$ represents $\bm{p}_a - \bm{p}_b$, similar to $\vec{V}_{bc}$ and $\vec{V}_{cd}$. 
The energy function is
\begin{equation}
    \mathcal{E}_{g}(\theta,\beta) = \|(\vec{V}_{ab} \times \vec{V}_{bc}) \cdot \vec{V}_{cd}\| - min(0, (\vec{V}_{ab} \times \vec{V}_{bc}) \cdot (\vec{V}_{bc} \times \vec{V}_{cd}))
\end{equation}

\noindent{\bf Identity Constraint.} We also have a regularization term to prevent the optimization from modifying the initialization too much. $\theta^\prime$ and $\beta^\prime$ are the initial values of $\theta$ and $\beta$.
\begin{equation}
    \mathcal{E}_{id}(\theta,\beta) = 
    \| \beta - \beta^\prime \|^2 + \| \theta - \theta^\prime \|^2
\end{equation}

\noindent{\bf Two-Step Optimization.} Eq. \ref{equ:scaleconstrain} and Eq. \ref{equ:jointconstrain} are of the same formulation, while their optimization objectives are different. We find that jointly optimizing the energy function is unstable since scale constraint is a global constraint while the location constraint is a local constraint. Therefore we adopt a two-step optimization scheme, optimizing hand scale first and hand details later, which accelerates the optimization convergence. 

\noindent{\bf Hand Scale Optimization.}
For $s^*$, Eq. \ref{equ:scaleconstrain} has an approximate analytical solution, we can obtain a near-optimal solution of $s^*$ in 1 iteration.
\begin{equation}\label{equ:s_persp}
    s^* = z_{root}\frac{\sum_{i=1}^k (f_ux_i, f_vy_i)(u_i-u_{root}, v_i-v_{root})^\top}{\sum_{i=1}^k |(f_ux_i, f_vy_i)|^2}
\end{equation}


\noindent{\bf Hand Detail Optimization.}
After obtaining a reasonable scale compensate factor, we can optimize the hand detail related energy function $\mathcal{E}$ in an iterative way. 
\begin{equation}\label{equ:energy}
    \mathcal{E}(\theta,\beta) = \lambda_J\mathcal{E}_{J}(\theta,\beta) + \lambda_{g}\mathcal{E}_{g}(\theta,\beta) + \lambda_{id}\mathcal{E}_{id}(\theta,\beta)
\end{equation}

The $\theta$ and the $\beta$ are updated with the gradient function of $\mathcal{E}$ by an optimizer in each iteration. The $\lambda$s are the coefficients for each term with corresponding subscripts.

\subsection{HandTailor In-The-Wild without $\mathcal{K}$}
\label{sec:in-the-wild}
The camera intrinsic $\mathcal{K}$ plays an important role in two critical points in the framework, transformation to camera coordinate (2D-3D mapping) and re-projection to 2D image plane (3D-2D mapping). However, sometimes $\mathcal{K}$ is unavailable, like an image from the internet, which limits the applicability of the framework. To address this HandTailor can be transited to weak perspective projection mode without any modification on network structure or weights.
Previous works for in-the-wild occasions usually treats the 2D-3D mapping by directly regressing 3D hand joints in camera coordinate \cite{zhou2020monocular} or regressing $\theta$ parameter \cite{boukhayma20193d,zhang2019end}, and the 3D-2D mapping by estimating the $s$ from neural networks \cite{boukhayma20193d,zhang2019end}. Such implicit treatments could cause interpretability and accuracy issues. Thanks to our multi-stage design, we find a way to calculate the scale factor $s$ in the weak-perspective projection analytically. It is noteworthy that though $s$ has the same meaning in both 2D-3D mapping and 3D-2D mapping, the estimation of $s$ should be treated differently since the information available is not the same in these two phases.

\noindent{\bf 2D-3D Mapping.}
In the 2D-3D mapping phase, we estimate the weak perspective projection scale factor $s$ (see Eq.\ref{equ:weakperspective}) by utilizing the reference bone prior, which is set to a unit length.
\begin{equation}
    [x_j,y_j,z_j]^\top = \Pi^{-1}(\mathcal{K}^\prime;[u_j^\prime,v_j^\prime,d_j^\prime]^\top)
\end{equation}
Consider the bone length $(\Delta x)^2+(\Delta y)^2+(\Delta z)^2=1$, we can calculate $s$ in $\mathcal{K}^\prime$ by
\begin{equation}
    s = \sqrt{\frac{\Delta u^{\prime2}+ \Delta v^{\prime2}}{1 - \Delta d^{\prime2}}}
\end{equation}
$[\Delta x, \Delta y, \Delta z]$ and $[\Delta u^\prime,\Delta v^\prime,\Delta d^\prime]$ are the reference bone vectors in different coordinate systems.

\noindent{\bf 3D-2D Mapping.} As for the 3D-2D mapping, we can rely on more plausible cues to estimate $s$, and along with the translation $t\in \mathbb{R}^2$ on the plane as follows:
\begin{equation}
   \mathcal{M}_{2D} = \Pi(\mathcal{M}*s) + t
\end{equation}
We can directly calculate them based on the 2D keypoints we predicted in the early stage. 
$t$ is the pixel coordinate of joint root $[u_{root},v_{root}] \in \mathcal{X}$. $s$ can be solved linearly by
\begin{equation}\label{equ:s_weak}
    s = \frac{\sum_{i=1}^k (x_i, y_i)(u_i-t_u, v_i - t_v)^\top}{\sum_{i=1}^k|(x_i,y_i)|^2}
\end{equation}

Then the same network can directly process images without any camera information by slightly changing the computation scheme.

\section{Experiments}
\subsection{Implementation} 

\noindent{\bf Hand Module.} The input resolution of \textit{hand module} is $256\times256$, and the intermediate resolutions are all $64\times64$. The training process is accomplished in a multi-step manner. We first train the keypoint estimation network for 100 epochs and IKNet for 50 epochs, finally train the whole network end-to-end for 100 epochs. We optimize the network through Adam with a learning rate of $3 \times 10^{-4}$ and decrease to $3 \times 10^{-5}$ in the end-to-end training stage. The network is trained with perspective projection. For in-the-wild occasions, we can directly change the computation scheme without any fine-tuning. The $\lambda$s from Eq. \ref{equ:losspose} are set to $100$, $1$ and $0.1$ respectively, and the $\lambda$s from Eq. \ref{equ:lossoverall} are set to $1$, $1$ and $0.1$ respectively.

\noindent{\bf Tailor Module.} The \textit{tailor module} is implemented on CPU with JAX \cite{jax2018github}, which can automatically differentiate the energy function, JIT compile and execute the optimization process. The scale-compensate factor is directly calculated via Eq. \ref{equ:s_persp}, and then optimize other constrains in an iterative manner to update $\beta$ and $\theta$. We utilize Adam with a learning rate of 0.003 as the optimizer, and the iteration number is set to $20$ for the trade-off between accuracy and time cost. The $\lambda$s from Eq. \ref{equ:energy} are set to $1$, $100$ and $0.1$ respectively.

\subsection{Experiment Setting}
\label{sec:expset}
We train and evaluate mainly on three datasets: Rendered Hand Dataset (RHD) \cite{zimmermann2017learning}, Stereo Hand Pose Tracking Benchmark (STB), and FreiHand dataset \cite{zimmermann2019freihand}. We report the percentage of correct keypoints (\textbf{PCK}), the area under the PCK curve (\textbf{AUC}), Procrustes aligned \cite{gower1975generalized} mean per joint position error
(\textbf{PA-MPJPE}), and Procrustes aligned mean per vertex position error (\textbf{PA-MPVPE}) as the main evaluation metrics. Note that previous works report AUC metric with a threshold range from $20mm$ to $50mm$, denoted as $AUC_{20-50}$. According to \cite{supancic2015depth}, it is because sometimes the annotation errors from real datasets can exceed $10mm$, and $20mm$ is agreeable to human judgment of two hands being close. However, as a synthetic dataset, RHD has no such problem. Thus to show the efficacy of \textit{tailor module}, we also report the AUC from $5mm$ to $20mm$, denoted as $AUC_{5-20}$ on the RHD dataset.

\subsection{Main Results}
\label{sec:mainresult}

\noindent{\bf Comparison with SOTA Methods.}
In Tab. \ref{tab:main}, we compare HandTailor with several previous state-of-the-art methods \cite{yang2020bihand,ge20193d,zhang2019end,boukhayma20193d,zimmermann2017learning,boukhayma20193d,kulon2020weakly,choi2020pose2mesh,moon2020i2l} on RHD, STB, and FreiHAND datasets. We can see that the proposed HandTailor can achieve state-of-the-art on RHD and STB benchmarks, and comparable results on FreiHAND with some mesh-convolution-based methods which involve extra mesh-level supervision, while HandTailor is a MANO-based method with only keypoint-level supervision. What's more, the architecture under weak perspective projection has only a little precision decay than perspective one, meaning that the weak perspective scheme can achieve reasonable simplification, and perspective one can have higher precision.

\begin{table}[h]
\begin{center}

\resizebox{0.9\linewidth}{!}{
\begin{tabular}{cc}
\begin{tabular}{c|c|c|c}
\toprule
\multirow{2}{*}{Method} 
& \multicolumn{2}{c|}{ RHD } & STB\\ 
\cline{2-4}
& $PCK_{20} \uparrow$ & \multicolumn{2}{c}{$AUC_{20-50} \uparrow$}  \\
\midrule\midrule
\cellcolor{Gray}Z\&B \cite{zimmermann2017learning} & \cellcolor{Gray}0.430 & \cellcolor{Gray}0.675 & \cellcolor{Gray}0.948 \\
Boukhayma et al. \cite{boukhayma20193d} & 0.790 & 0.926 & 0.995 \\
\cellcolor{Gray}Zhang et al. \cite{zhang2019end} & \cellcolor{Gray}0.740 & \cellcolor{Gray}0.901 & \cellcolor{Gray}0.995 \\
Ge et al.\cite{ge20193d} & 0.810 & 0.920 & \textbf{0.998} \\
\cellcolor{Gray}Yang et al.\cite{yang2020bihand} & \cellcolor{Gray}0.846 & \cellcolor{Gray}0.951 & \cellcolor{Gray}0.997 \\
\midrule
\textit{Hand module} & 0.833 & 0.949 & 0.997 \\
\cellcolor{Gray}HandTailor & \cellcolor{Gray}\textbf{0.874} & \cellcolor{Gray}\textbf{0.958} & \cellcolor{Gray}\textbf{0.998} \\
HandTailor (w/o $\mathcal{K}$) & 0.829 & 0.932 & 0.991 \\
\bottomrule
\end{tabular}
&
\begin{tabular}{c|c|c}
\toprule
\multirow{2}{*}{Method} 
& \multicolumn{2}{c}{FreiHAND}\\ 
\cline{2-3}
& $PA-MPJPE \downarrow$ & $PA-MPVPE \downarrow$\\
\midrule\midrule
\cellcolor{Gray}Boukhayma et al. \cite{boukhayma20193d} & \cellcolor{Gray}35.0 & \cellcolor{Gray}13.2  \\
MANO CNN \cite{zimmermann2019freihand} & 11.0 & 10.9  \\
\cellcolor{Gray}YoutubeHand \cite{kulon2020weakly} & \cellcolor{Gray}8.4 & \cellcolor{Gray}8.6  \\
Pose2Mesh \cite{choi2020pose2mesh} & 7.7 & 7.8  \\
\cellcolor{Gray}I2L-MeshNet \cite{moon2020i2l} & \cellcolor{Gray}\textbf{7.4} & \cellcolor{Gray}\textbf{7.6}  \\
\midrule
\textit{Hand module} & 8.5 & 8.8  \\
\cellcolor{Gray}HandTailor & \cellcolor{Gray}8.2 & \cellcolor{Gray}8.7\\
HandTailor (w/o $\mathcal{K}$) & 8.9 & 9.2\\
\bottomrule
\end{tabular}
\\
\end{tabular}
}
\end{center}
\caption{Comparison with state-of-the-art methods}
\label{tab:main}
\end{table}

\noindent{\bf Efficacy of Tailor Module.} To better reflect the influence of \textit{tailor module}, we select stricter metrics, PCK ranges from $5mm$ to $20mm$. Also as we mentioned before, the \textit{tailor module} only relates to the intermediate and final results provided by networks, so we conduct the plug-and-play experiments on several existing methods \cite{yang2020bihand,boukhayma20193d,zhou2020monocular} with the demo models released by authors. 

\begin{table}[h]
\vspace{-3mm}
\begin{center}
\resizebox{0.45\linewidth}{!}{
\begin{tabular}{c|cccc}
\toprule
Method & Ours & \cite{yang2020bihand} & \cite{boukhayma20193d} & \cite{zhou2020monocular} \\
\midrule\midrule
\cellcolor{Gray}Original & \cellcolor{Gray}0.561 & \cellcolor{Gray}0.568 & \cellcolor{Gray}0.274 & \cellcolor{Gray}0.314 \\
+ \textit{Tailor module} & \textbf{0.658} & \textbf{0.601} & \textbf{0.341} & \textbf{0.356} \\
\bottomrule
\end{tabular}
}
\end{center}
\caption{A stricter metric $AUC_{5-20}$ of previous work and the proposed HandTailor on RHD.}
\label{tab:main2}
\vspace{-3mm}
\end{table}

\begin{figure*}[t]
   \begin{center}
   \includegraphics[width=0.85\linewidth]{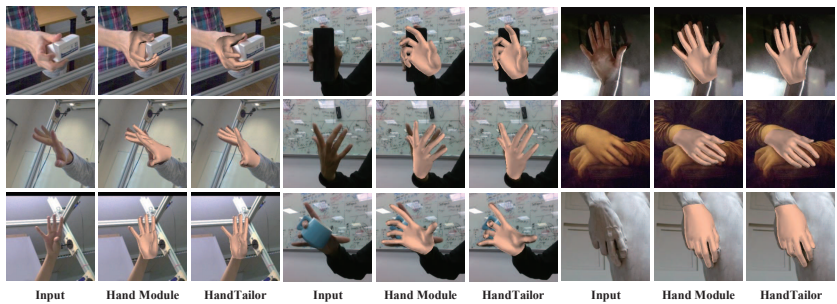}
   \end{center}
   \vspace{-5mm}
   \caption{Qualitative result of HandTailor. There are three examples in each row. For each example, there are input image (left), output of \textit{hand module} (middle), and the result of HandTailor (right). We can see that the \textit{tailor module} improve the quality of the hand mesh remarkably. Our demonstrations include the samples from FreiHAND dataset \cite{zimmermann2019freihand}, images captured by RealSense D435, and some pictures downloaded from the internet. The samples from FreiHAND and the internet are evaluated using weak-perspective projection, while samples from RealSense are evaluated using perspective projection.}
   \label{fig:quantitative}
\vspace{-4mm}
\end{figure*}

\begin{figure}[h]
\begin{center}
\vspace{-3mm}
\subfigure[]{
\centering
\includegraphics[width=0.258\linewidth]{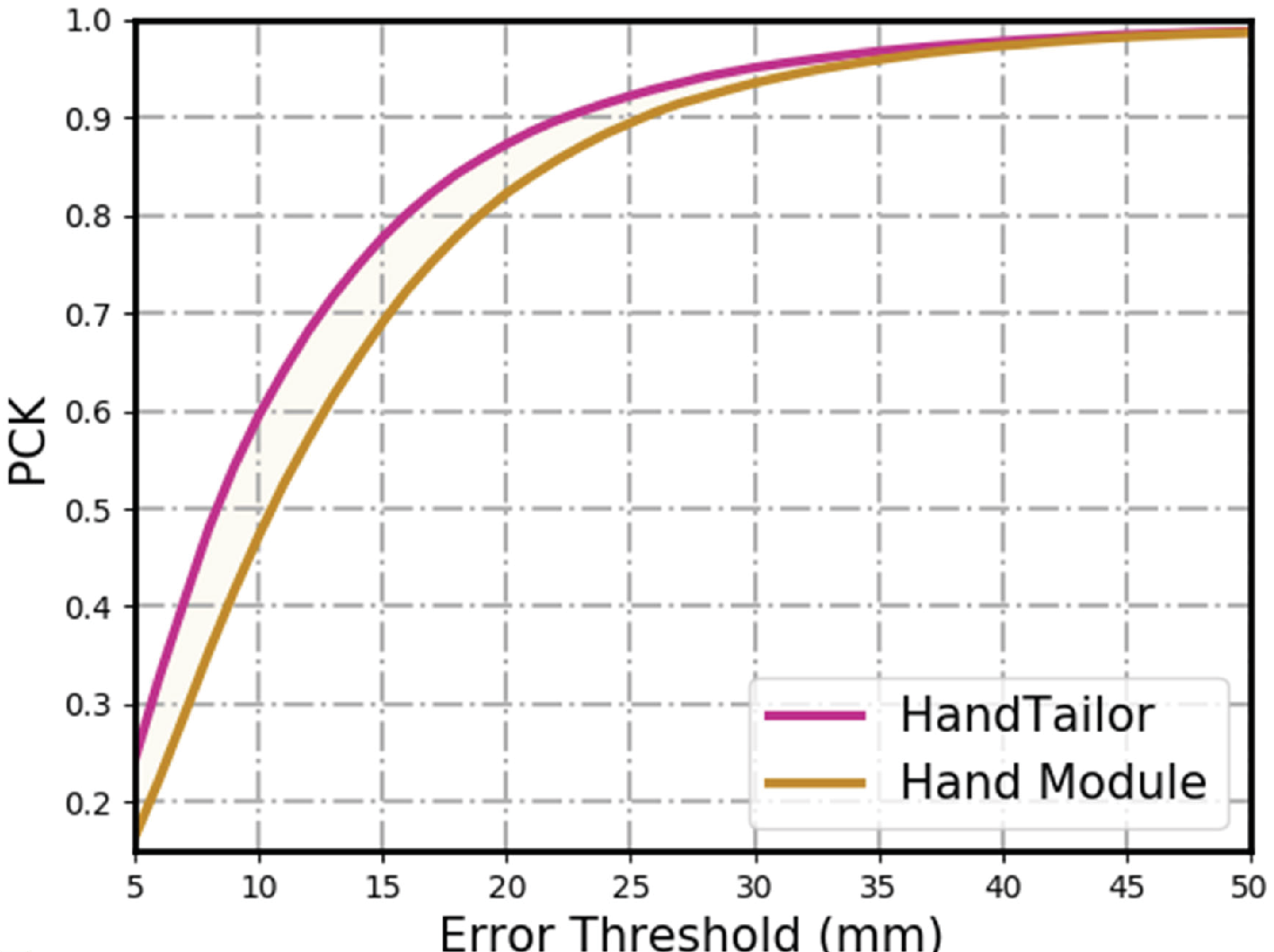}
\label{fig:auc}
}%
\subfigure[]{
\centering
\includegraphics[width=0.25\linewidth]{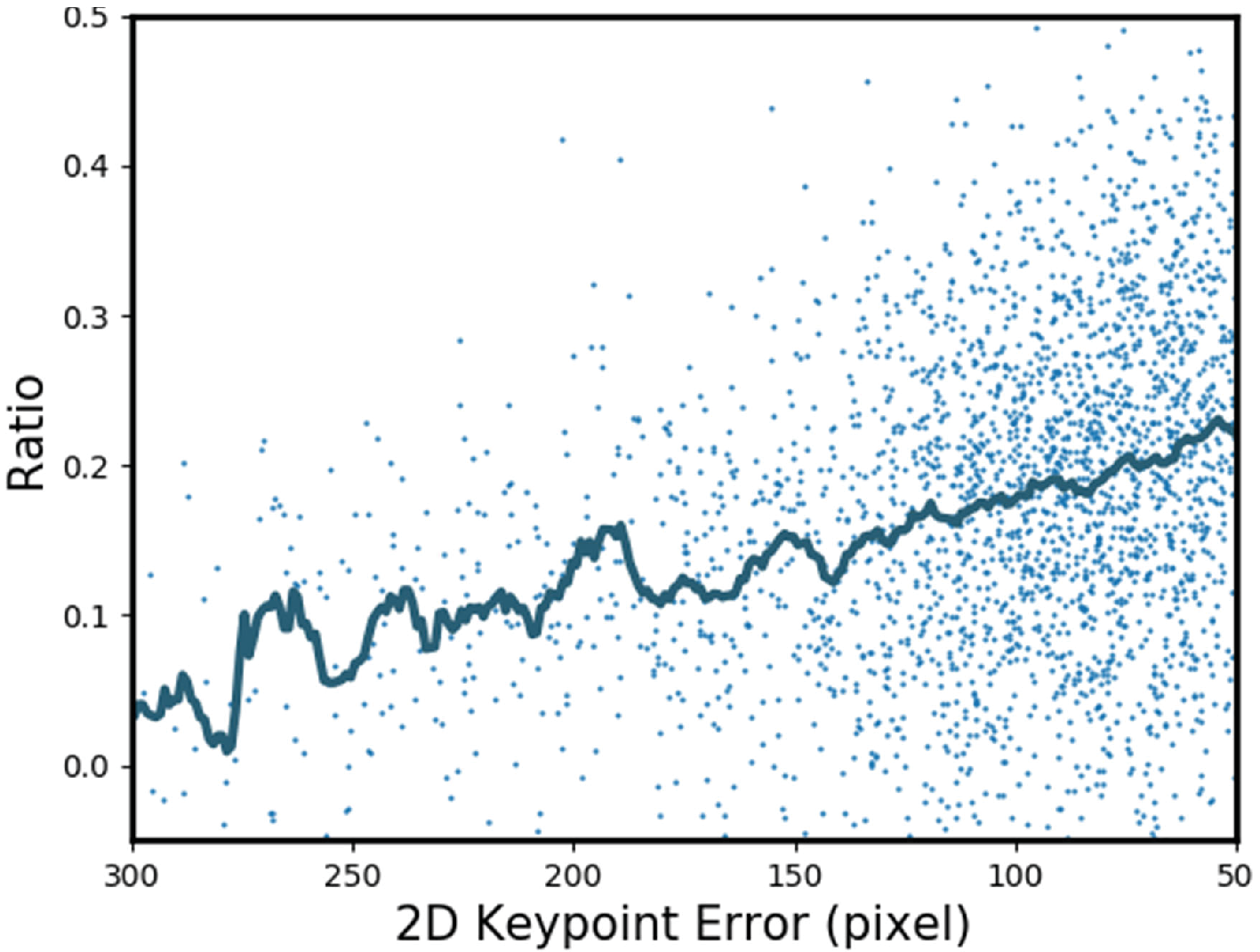}
\label{fig:tailorcurve}
}
\end{center}
\vspace{-4mm}
\caption{(a): PCK curve with a threshold ranges from $5mm$ to $50mm$ on RHD. (b): The abscissa represents the error of 2D keypoint estimation of a specific sample on RHD, and the ordinate represents the influence of \textit{tailor module}, which is the ratio of 3D joints location error of \textit{hand module} and the error eliminated by the \textit{tailor module}.}
\end{figure}

As shown in Tab. \ref{tab:main2} and Fig. \ref{fig:auc}, the \textit{tailor module} can bring a significant improvement, which proves its effectiveness in reducing errors and suiting many different networks. To show how the quality of intermediate results influences the performance of \textit{tailor module}, we test and record the error of 2D keypoints and 3D joints of HandTailor on every sample in RHD. In Fig. \ref{fig:tailorcurve}, each point denote a sample in RHD dataset. The abscissa represents the 2D keypoint error of the sample, and the ordinate represents the ratio of 3D joints location error of hand module and the error eliminated by the tailor module. We can find that when the 2D keypoint estimation has higher precision, the \textit{tailor module} can play a bigger role.

\noindent{\bf Speed of Tailor Module.} Optimization-based methods tend to be slow. But the proposed \textit{tailor module} can achieve high accuracy with a tiny time cost.
We conduct the experiments with $20$ iterations, which cost only $8.02ms$ per sample on PC with Intel i7-8700 CPU. 

\begin{table}[h]
\begin{center}
\resizebox{0.4\linewidth}{!}{
\begin{tabular}{c|cccc}
\toprule
Iteration & 10 & 20 & 50 & 100 \\
\midrule\midrule
Time ($ms$) & 4.27 & 8.02 & 19.21 & 39.53 \\
\cellcolor{Gray}$AUC_{20-50}$ & \cellcolor{Gray}0.952 & \cellcolor{Gray}0.956 & \cellcolor{Gray}0.956 & \cellcolor{Gray}0.956 \\
$AUC_{5-20}$ & 0.624 & 0.653 & 0.662 & 0.666 \\
\bottomrule
\end{tabular}
}
\end{center}
\caption{The time cost and accuracy on RHD with different \textit{tailor module} iteration numbers.}
\label{tab:speed}
\end{table}

\noindent{\bf Qualitatively Result.} It can be seen from Fig. \ref{fig:quantitative} that when we only rely on the \textit{hand module}, the hand skeleton seems correct, but once it is re-projected onto the original image, there will be numerous evident small errors, such as a few degrees of finger deviation. Once we utilize the \textit{tailor module} to fine-tune the hand reconstruction results, these errors can be largely fixed, and we can make the projection more coherent to the image. 

\subsection{Ablation Study} 
\label{sec:ablation}
In this part, we evaluate some key components of our approach on RHD.


\noindent{\bf Losses of Shape-Branch.} To supervise the shape-branch of \textit{hand module}, despite the regularization loss $\mathcal{L}_{\beta}$ mentioned before, a depth loss $\mathcal{L}_{\beta^{D}}$ and a silhouette loss $\mathcal{L}_{\beta^{S}}$ can also be conducted through a differential renderer \cite{kato2018neural}. $\mathcal{L}_{\beta^{D}}$ and $\mathcal{L}_{\beta^{S}}$ are the MSE losses between predicted and ground-truth depth and silhouette. Tab. \ref{tab:ab} shows that these losses cannot enhance the performance effectively. So we only use a simple regularization $\mathcal{L}_{\beta}$ while training.


\noindent{\bf Energy Function.} The energy function of \textit{tailor module} has three components affecting the performance on PCK metric: $\mathcal{E}_{J}$, $\mathcal{E}_{g}$, and $\mathcal{E}_{id}$.
Besides these terms, the \textit{tailor module} can also utilize depth constraint $\mathcal{E}_{d}$ that measures the distance between rendered and input depth, and silhouette constraint $\mathcal{E}_{s}$ that measures the distance between rendered and input silhouette. The rendered depth and silhouette are generated via a differential renderer, and the input depth and silhouette are both extracted by watershed from the depth map.

\begin{table}[h]
\begin{center}
\resizebox{0.8\linewidth}{!}{
\begin{tabular}{cc}
\begin{tabular}{ccc|cc}
\toprule
$\mathcal{L}_{\beta}$ & $\mathcal{L}_{\beta^{D}}$ & $\mathcal{L}_{\beta^{S}}$ & $AUC_{20-50}$ & $AUC_{5-20}$ \\
\midrule\midrule
 \checkmark &  &   & 0.958& 0.658\\
 \cellcolor{Gray}\checkmark & \cellcolor{Gray}\checkmark & \cellcolor{Gray}  & \cellcolor{Gray}0.956 & \cellcolor{Gray}0.653\\
 \checkmark &  & \checkmark  & 0.955& 0.651\\
\bottomrule
\end{tabular} &
\begin{tabular}{cccc|ccc}
\toprule
$\mathcal{E}_{g}$ & $\mathcal{E}_{id}$ & $\mathcal{E}_{d}$ & $\mathcal{E}_{s}$ & $AUC_{20-50}$ & $AUC_{5-20}$ & Time($ms$)\\
\midrule
\midrule
 \checkmark & \checkmark &  &  & 0.958 & 0.658 & 8.02\\
 \cellcolor{Gray}\checkmark & \cellcolor{Gray} & \cellcolor{Gray} & \cellcolor{Gray} & \cellcolor{Gray}0.958 & \cellcolor{Gray}0.657 & \cellcolor{Gray}8.01\\
   & \checkmark &  &  & 0.958 & 0.658 & 7.98\\
 \cellcolor{Gray}\checkmark & \cellcolor{Gray}\checkmark & \cellcolor{Gray}\checkmark & \cellcolor{Gray} & \cellcolor{Gray}0.950 & \cellcolor{Gray}0.644 & \cellcolor{Gray}42.11\\
 \checkmark & \checkmark &  & \checkmark & 0.948 & 0.640 & 42.08\\
\bottomrule
\end{tabular} \\
\end{tabular}
}
\end{center}
\caption{Ablation study on the loss functions of shape-branch (left), and how different constraints of \textit{tailor module}. influence the precision and speed (right).}
\label{tab:ab}
\end{table}

\begin{figure}[h]
\begin{center}
\includegraphics[width=0.4\linewidth]{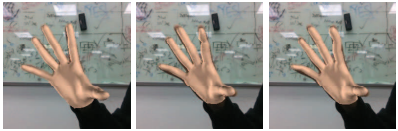}
\end{center}
\vspace{-5mm}
  \caption{From left to right are the output of the full \textit{tailor module}, the output of the \textit{tailor module} without unnatural twist constraint, and the output without identity constraint.}
\label{fig:energy}
\vspace{-5mm}
\end{figure}

We can see from Tab. \ref{tab:ab} that 
$\mathcal{E}_{g}$ and $\mathcal{E}_{id}$ do not improve the precision, but they do affect the hand mesh quality as shown in Fig. \ref{fig:energy}. Also, depth and silhouette constraints cannot improve \textit{tailor module}. This is because they make the energy function more complex and hard to converge. These constraints also bring a heavy overhead to \textit{tailor module}.

\section{Conclusion}

In this paper, we propose a novel framework HandTailor for monocular RGB 3D hand recovery, combining a learning-based \textit{hand module} and an optimization-based \textit{tailor module}. We can adapt both perspective projection and weak perspective projection for high precision and in-the-wild scenarios. And the \textit{tailor module} can bring significant improvement for the whole pipeline both qualitatively and quantitatively. In the future, we will try to solve the hand reconstruction task when the hand is holding objects or two hands are interacting.

\bibliography{egbib}
\end{document}